\pdfoutput=1

\documentclass[runningheads]{llncs}

\usepackage{graphicx}

\input{psfig.sty}

\begin{document}

\pagestyle{headings}

\mainmatter

\title{3D Face Pose and Animation Tracking via Eigen-Decomposition based Bayesian Approach}

\author{Ngoc-Trung Tran\inst{1}, Fakhr-Eddine Ababsa\inst{2},  Maurice Charbit\inst{1}, Jacques Feldmar\inst{1} , Dijana Petrovska-Delacr{\'e}taz\inst{3} and G{\'e}rard Chollet\inst{1}}
\institute{ ENST, 75014 Paris, France\inst{1} \\ 
           	\email{\{trung-ngoc.tran,maurice.charbit,gerard.chollet\}@enst.fr} \\ 
           	\email{jfeldmar@gmail.com} 
           	\and IBISC, 91020 Evry, France\inst{2} \\
           	\email{ababsa@iup.univ-evry.fr} \\
           	Telecom Sudparis, 91000 Evry, France\inst{3}\\ 
           	\email{dijana.petrovska@telecom-sudparis.eu}
           }

\titlerunning{Lecture Notes in Computer Science}

\maketitle

\begin{abstract}
This paper presents a new method to track both the face pose and the face
animation with a monocular camera. The approach is based on the 3D face model CANDIDE and on the SIFT (Scale Invariant Feature Transform) descriptors, extracted around a few given landmarks (26 selected vertices of CANDIDE model) with a Bayesian approach. The training phase is performed on a synthetic database generated from the first video frame. At each current frame, the face pose and animation parameters are estimated via a Bayesian approach, with a Gaussian prior and a Gaussian likelihood function whose the mean and the covariance matrix eigenvalues are updated from the previous frame using eigen decomposition. Numerical results on pose estimation and landmark locations are reported using the Boston University Face Tracking (BUFT) database and Talking Face video. They show that our approach, compared to six other published algorithms, provides a very good compromise and presents a promising perspective due to the good results in terms of landmark localization.
\end{abstract}

\section{Introduction}

Tracking 3D face pose is an important issue and has received much attention in the last decades because of \textit{multiple applications} involved such as: video surveillance, human computer interface, biometrics, \textit{etc}. And it is much more challenging if the face animation or expression needs to be recognized in the meantime in variety of applications. Difficulties come from a number of factors such as projection, multi-source lighting biological appearance variations, facial expressions as well as occlusions with accessories, \textit{e.g.,} glasses, hats... In this paper, we present a method using the model of landmarks to track pose efficiently as well as model facial animation. Note that the face is controlled by shape and animation which could be validated as landmark tracking problem. 

Since the pioneer work of \cite{cootes-bmvc-1992,cootes-ieeepami-2000}, it is well-known that the Active Shape Model (ASM) and Active Appearance Model (AAM) provide an efficient approach for face pose estimation and tracking landmarks of frontal or near-frontal faces. Some extensions \cite{xiao-cvpr-2004,gross_ivc_2006} have been developed to improve the method in terms of accurate landmarks or profile-view fitting. Recently, Saragih \textit{et al.} \cite{saragih-ijcv-2011} via exhaustive local search around landmarks constrained by a 3D shape model, can track single face of large Pan angle in well-controlled environment. However, it needs a lot of annotated data, which is costly in unconstrained environments, to learn 3D shape and local appearance distributions. One another approach tracks faces and estimate pose uses 3D rigid models such as semi-spherical or cylinder \cite{cascia-ieeepami-2000,xiao-ijist-2003}, ellipsoid \cite{morency-fg-2008} or mesh \cite{vacchetti-ieeepami-2004}. These methods can estimate three rotations well even profile-view; however, non-rigid transformation can not be applied for animation problem.

For those who using synthesized databases or online tracking technique with 3D face. An early proposal \cite{decarlo-ijcv-2000} concerns optical flow and does adaptable changes. Optical flow can be very accurate but not robust on fast movements. Moreover, this approach accumulates errors to drift away and is not easy to recover in long video sequences. With the help of local features, which provides invariant descriptors to non-rigid motions, Chen and Davoine \cite{chen-bmvc-2006} took advantages of local features constrained by a 3d-face paramerized model, called Candide-3, to capture both rigid and non-rigid head motions. But this methods does not work well in profile-view due to the large variation of landmarks. Ybanez \textit{et al.} \cite{alonso-icip-2007} found linear correlation between 3D model parameters and global appearance of stabilized face images. This method is robust for face and landmark tracking but limited just around frontal faces. Lefevre \textit{et al.} \cite{lefevre-icme-2009} extended Candide by collecting more appearance information at profile-views and chose more random points to represent facial appearance. Their error function consists of structure and appearance features combined with dynamic modeling, is high dimension and is easy to fall into local minimum. Tran \textit{et al.} \cite{tran-visapp-2013} uses the sparse representation to formalize the objective for 3d face tracking. The codebook of patches is constructed from the synthesized dataset. Recently, faceAPI \cite{faceapi} showed impressive results in pose and face animation tracking; however, this is a commercial product that unable to be accessed to investigate and compare with other methods.

In this paper, we propose an Bayesian method using a 3D face model to build the face pose and animation tracking framework. Our contribution is that in our framework, the SIFT \cite{lowe-ijcv-2004} is supposed to be local descriptor to track landmarks which are constrained by the 3D shape. And eigen decomposition is proprosed to use through Singular Value Decomposition (SVD) to update the tracking model robustly and balance between what we learned in training and what we are seeing at the moment. This approach is different what previous methods of face tracking did. We also take advantages of a synthesized database \cite{chen-bmvc-2006,alonso-icip-2007,lefevre-icme-2009} without the need of big annotated data and propose the use of robust features to rigid and non-rigid changes. During tracking, candidate of new pose and animation is estimated via the posterior probability and the appearance model are then adjusted from new observations to environmental changes. This technique can make the system robust to changes of facial expression, pose and as well as environmental factors. The results on two public datasets show that our approach, compared to six other published algorithms, provides a very good compromise in terms of pose estimation and landmark localization.

The remaining of this paper are organized as follow: Section 2 gives some background face representation. Section 3 shows the proposed framework for tracking. Experimental results and analysis are presented in Section 4. Finally, we draw conclusions in Section 5.

\section{Face Representation}

\begin{figure*}
\begin{centering}
\includegraphics[scale=0.35]{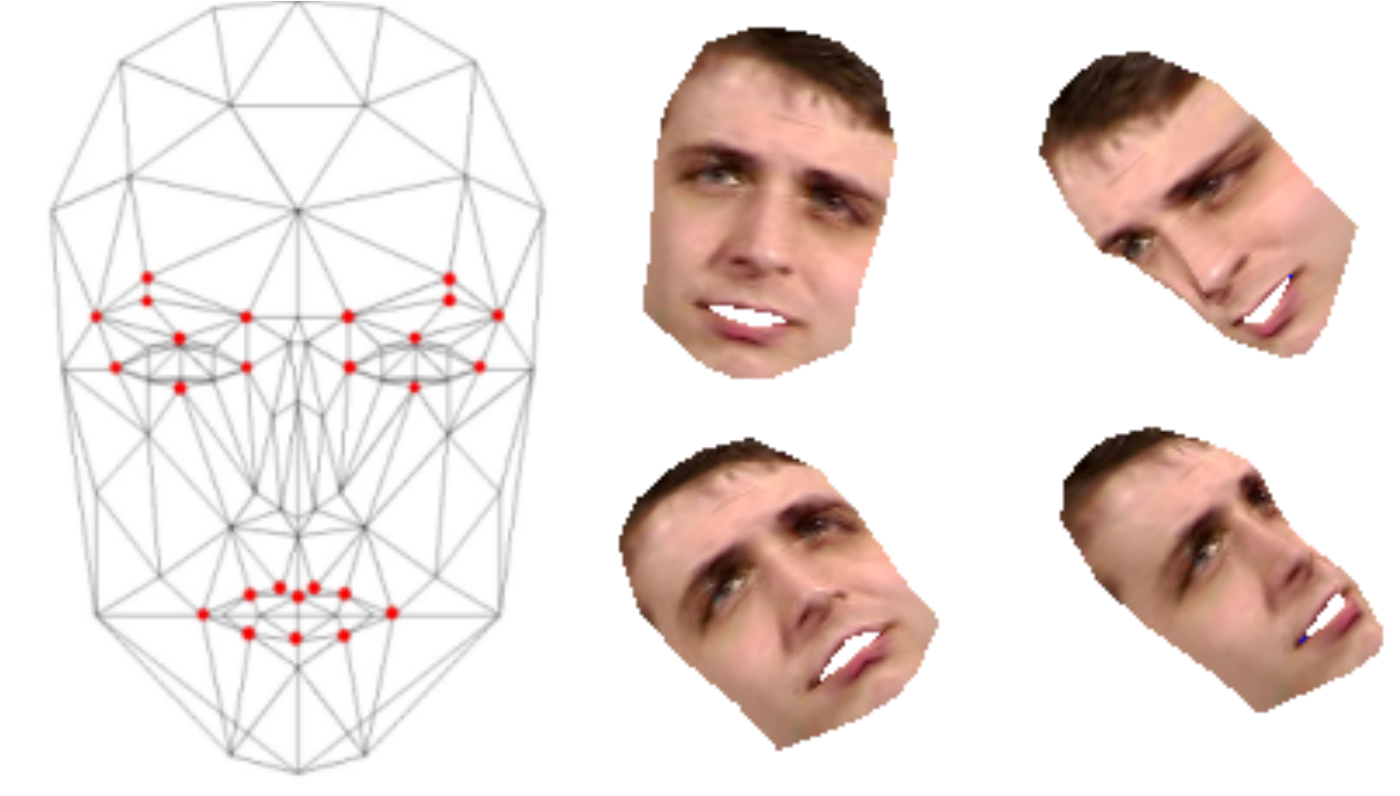}
\caption{Candide-3 and some sample synthesized images.}
\end{centering}
\end{figure*}


Candide-3 \cite{ahlberg-tech-2001} is a very commonly used face shape model. It consists of 113 vertices and 168 surfaces. Fig. 1 represents the frontal view of the model. It is controlled both in translation, rotation, shape and animation:

\begin{equation}
\textrm{g}(\sigma,\alpha)=\textrm{R}s\left(\overline{\textrm{g}}+\textrm{S}\sigma+\textrm{A}\alpha\right)+\textrm{t}\label{eq:1}
\end{equation}

\noindent where $\overline{\textrm{g}}$ is 3N-dimensional mean shape (N = 113 is
the number of vertices) containing the 3D coordinates of the
vertices. The matrices S and A control respectively shape and animation
through $\sigma$ and $\alpha$ parameters. $\textrm{R}$ is a rotation matrix, s is the scale, and $\textrm{t}$ is the translation
vector. The model makes an perspective projection assumption to project 3D face onto 2D image. Like \cite{chen-bmvc-2006,lefevre-icme-2009,alonso-icip-2007}, only 6 dimensions $r_{a}$ of the animation parameter are used to track eyebrows, 
eyes and lips. Therefore, the full model parameter b of our framework has 12 dimensions: of 3 dimensions for rotation $(r_{x},r_{y},r_{z})$, 3 dimensions for translation $(t_{x},t_{y},t_{z})$ and 6 dimensions for animation $r_{a}$:

\begin{equation}
\textrm{b}=[r_{x},r_{y},r_{z},t_{x},t_{y},t_{z},r_{a}]\label{eq:2}
\end{equation}

\textbf{Texture model:} In the Candide model, appearance or texture parameters are not 
available. Usually, we warp and map the image texture onto the triangles
of the 3d mesh by the image projection.

\section{Proposed Method}
\label{sect:figures}

Our framework consists of two steps: training and tracking. The framework benefits a database of synthesized faces to train tracking model and applies new way of tracking face pose and animation. In this section, we describe our method in detail.

\subsection{Training}

In the work of \cite{chen-bmvc-2006}, the authors align manually the Candide model on the first video frame and warp and map the texture from the image to the model. In our work, landmarks are annotated manually on the first video frame, then the POSIT algorithm \cite{dementhon-ijcv-1995} is used to fit and estimate the pose automatically from these landmarks to get the initial model parameters $b_{0}$.

The acquisition of ground-truth is very costly and time consuming. In order to circumvent this drawback, synthetic database \cite{chen-bmvc-2006,alonso-icip-2007,lefevre-icme-2009} using the Candide model is a good alternative. In order to collect training data, we do three following steps to obtain images using Candide and build appearance model for the next tracking step:

\subsubsection{Data Generation}

After initialization, the texture is warped and mapped from the first video frame to the
Candide model. Our database is built by rendering different views around the frontal image. Note that the full dimension of 
the parameters to track is 12, consists of pose and animation, that makes difficult to explore finely. However, the translation parameters $t_{x}$ and $t_{y}$
will not affect the face appearances as well as facial animation will not be significant
influence because the use of local features in tracking. Hence, only rotations are gridded
for building the training database. Specifically, 7 values of Pan and Tilt and Roll 
from -30 to +30 by step of 10 are taken to create $7^{3}=343$ pose views as some examples in Fig.
1.

\subsubsection{Learning Appearance Model}

The framework adopts local descriptors which are robust to rigid and non-rigid motion. In this paper, we also use SIFT descriptor \cite{lowe-ijcv-2004} to extract local features around 26 given landmarks in Fig. 1 as observed appearance. SIFT is invariant to affine transformation and helpful to localize accurate landmarks. In order to get the appearance model, we compute mean and covariance matrices of landmark descriptors on 343 images of the synthesized database which is generated from the first image. Each pair of mean and covariance matrix $(\mu^{i},\Sigma^{i})$ plays the role of learning data for ith landmark which are  $128\times1$ and $128\times128$ matrices respectively. And these matrices will be adjusted during tracking.

\subsection{Tracking}


Here we propose a Bayesian approach approximated from posteriori distribution:

\begin{equation}
p(b_{t}|Y_{1:t})=\frac{p(Y_{t}|b_{t},Y_{1:t-1})p(b_{t}|Y_{1:t-1})}{p(Y_{t}|Y_{1:t-1})}\propto p(Y_{t}|b_{t},Y_{1:t-1})p(b_{t}|Y_{1:t-1}) \label{eq:3}
\end{equation}

Equation (3) is normally controlled by the observation model $p(Y_{t}|b_{t},Y_{1:t-1})$, and the evolution $p(b_{t}|Y_{1:t-1})$ as the prior. Because Eq. \ref{eq:3} is still complicated to solve, we provide some assumptions to make it simpler. 

\subsubsection{Evolution Model}

The model $p(b_{t}|Y_{1:t-1})$ of state $b_{t}$ is dependent on only previous observation $Y_{1:t-1}$. We know $\hat{b}_{t-1}$ was able to estimated from $Y_{1:t-1}$. So, we assume that $p(b_{t}|Y_{1:t-1}) \propto p(b_{t}|\hat{b}_{t-1})$ which means $b_t$ is modeled independently by a Gaussian distribution around its previous estimated state $\hat{b}_{t-1}$, where $b_t=(r_{x},r_{y},r_{z},t_{x},t_{y},t_{z},r_{a})_t$ is the 12-dimensional vector in our context expressed as:

\begin{equation}
p(b_{t}|\hat{b}_{t-1})=\mathcal{N}(b_{t};\hat{b}_{t-1},\Psi) \label{eq:4}
\end{equation}

\noindent where $\Psi$ is a diagonal covariance matrix whose elements are the corresponding variances
of parameters of the state vector $\sigma^{i},i=1,..,12$. This model can be considered as the prior information during tracking.

\subsubsection{Observation Model}

The tracking system starts from the frontal face where Candide is fitted onto, and then it finds the candidate 
of face in the next frame $t+1$ from the state vector at time $t$, with $t=0$ at the first frame.
In order to obtain the observation $Y_{t}$, the 3d Candide model is projected onto the next 2D frame at $t$ to localize 2D landmark positions. The appearance $Y_{t}$ is a vector of local textures $(y^{1}_{t}, y^{2}_t, ...,y^{n}_{t})$ around these landmarks as the observation. These observations can then be used to establish the observation model for tracking and the crucial point is to find an efficient observation model.

We make the assumption that the local appearances around landmarks are independent. The observation model is defined as a joint probability of Gaussian distributions, and the tracking problem can be solved as a maximum likelihood problem of a non-linear function.

\begin{equation}
p(Y_{t}|b_{t},Y_{1:t-1}) = {\prod}_{i=1}^{n} p(y^i_{t}|b_{t},y^i_{1:t-1}) \label{eq:5}
\end{equation}

It means that the observation  $Y_{t}$ is dependent on the state variable $b_{t}$ as well as previous observations $Y_{t-1}$. Since the database of synthesized faces is generated in the range limit of $(-30;30)$ of three rotations that make the system limited in profile tracking. We can generate more data, however, it makes the framework less robust because of the variation for patches as well as occlusion problem at profile-view. Additionally, there are many factors such as illumination, poses and facial expression that may affect to tracking. So, the learning model needs to be adaptive to changes of environment that brings us the idea of maximum likelihood problem (\ref{eq:5}) can be rewritten as follows:

\begin{equation}
p(Y_{t}|b_{t},Y_{1:t-1})={\prod}_{i=1}^{n}\mathcal{N}(y^{i}_{t}|\mu^{i}_{t},\Sigma^{i}_{t})\label{eq:6}
\end{equation}

\noindent where n is the number of landmarks, $\mathcal{N}(y^{i}_{t}|\mu^{i}_t,\Sigma^{i}_t)$ denotes multivariate Gaussian distribution
of function value at observation around the ith landmark $y^{i}_{t}$, and $\mu^{i}_{t}$ and $\Sigma^{i}_{t}$ are mean and covariance matrices updated at time $t$ during tracking. Note that $\mu^{i}_{0}$ and $\Sigma^{i}_{0}$ are pre-learned mean and covariance in training step at first frame. The likelihood in Eq. \ref{eq:6} is controlled by two terms: $\mu^{i}_{t}$ and $\Sigma^{i}_{t}$ which model how confidence the new landmark observation is. Since trained at first frame, these terms should be adjusted to fit changes of factors, but still "remember" what it learned before. The proposed way how to update can be described as follows for mean vectors:

\begin{equation}
\mu^{i}_{t} = (1-\alpha)\mu^{i}_{t-1} + \alpha y^{i}_{t-1}\label{eq:7}
\end{equation}

\noindent where forgetting factor $\alpha \in (0,1)$ is a constant. This equation is a way to correct the error between the observation and the mean vector of appearance model. In order to update covariance matrices, Singular Value Decomposition (SVD) \cite{golub-siam-1965} is used to factorize the previous covariance matrix at time $t-1$ into unitary matrices and singular matrix of eigen values: $svd(\Sigma^{i}_{t-1})=[U^{i}_{t-1},S^{i}_{t-1},(U^{i}_{t-1})^T]$. Note that covariance matrix is positive definite, so unitary matrices are the same. Then, updating the singular matrix before composing all of them back to obtain a new covariance matrix at time $t$.

\begin{equation}
S^{i}_{t}=(1-\alpha)S_{t-1}^{i}+\alpha \left\Vert y^{i}_{t-1}-\mu^{i}_{t-1} \right\Vert ^{2}_2I
\quad \texttt{and} \quad
\Sigma^{i}_{t} = U^{i}_{t-1}S_{t}^{i}(U^{i}_{t-1})^{T}
\label{eq:8}
\end{equation}

\noindent where $I$ is identity matrix, $\left\Vert.\right\Vert_{2}$ is norm-2. The equations denote how to do adaptive observation model, while keeping principal components of what is seen before. In order to do this, we use Eq. \ref{eq:7} for the eccentricity and the direction is changed when the new covariance matrix is decomposed to update in next step as Eq. \ref{eq:8}. The updated mean and covariance matrices are used to model the observation as Eq. \ref{eq:6}. To sum up, replacing the observation and evolution models respectively of equations (\ref{eq:4}) and (\ref{eq:6}) into (\ref{eq:3}) and taking the log of likelihood, we finally attempt to minimize the error function approximated as follows:

\begin{equation}
\hat{b}_{t}=\arg\min_{b_{t}}\sum_{i=1}^{n}\left\Vert y^{i}_{t}-\mu^{i}_{t}\right\Vert^{2}_{(\Sigma^{i}_{t})^{-1}} + \left\Vert b_{t} - \hat{b}_{t-1} \right\Vert_{\Psi^{-1}}^{2}\label{eq:9}
\end{equation}

\noindent where $\hat{b}_{t-1}$ is the model parameter estimated from previous frame. In our optimization context, the error function in (\ref{eq:9}) is a multi-dimensional function of the model parameter $b_{t}$ that we wish to minimize. It is not easy
to solve analytically, so a derivative-free optimizer such as down-hill
simplex \cite{nelder-cj-1965} is preferred. Like \cite{chen-bmvc-2006},
thirteen initial points are chosen randomly around the current state (12-dimensional
space) to form the simplex and the solution that subjects to local minimum can be found by deformations
and contracts during optimization.

\section{Experimental Results}
\label{sect:TeX}

\begin{figure*}
\begin{centering}
\includegraphics[scale=0.60]{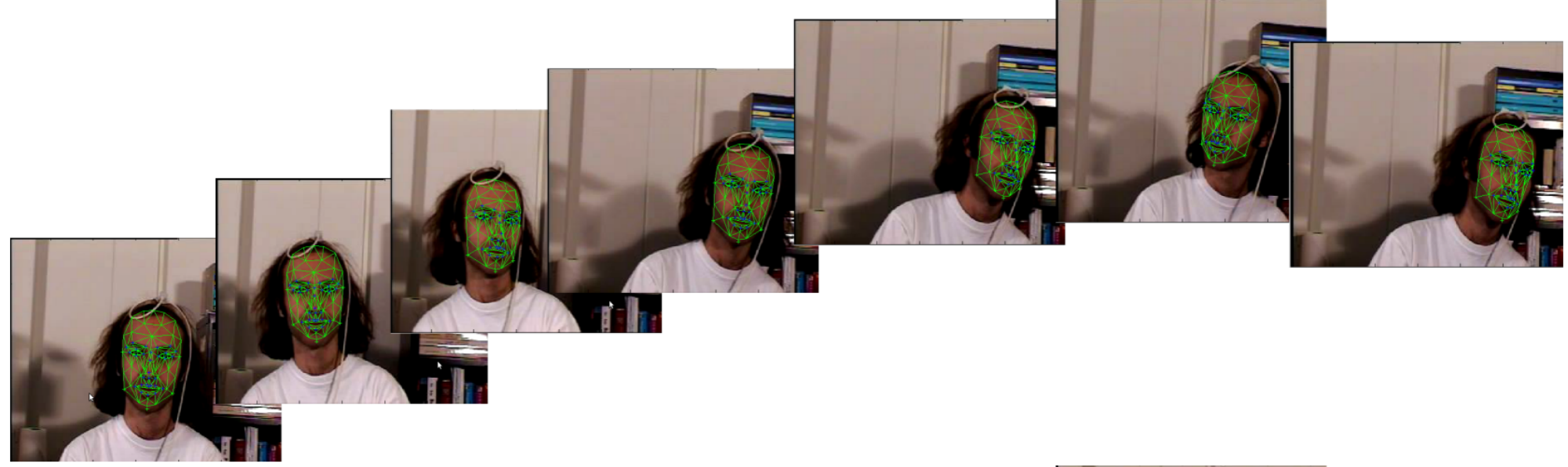}
\caption{An sample result of our method on one BUFT video.}
\end{centering}
\end{figure*}


We adopted the Boston
University Face Tracking (BUFT) database \cite{cascia-ieeepami-2000} and Talking Face video\footnote{http://www-prima.inrialpes.fr/FGnet/data/01-TalkingFace/talking\_face.html} to evaluate the performances of face pose estimation and its animation by landmark tracking respectively.

\textbf{BUFT:} The pose ground-truth is captured by magnetic sensors {}``\textit{Flock
and Birds}'' with an accuracy of less than $1^{o}$. The uniform-light set which is used to evaluate, has a total 
of 45 video sequences (320$\times$240 resolution) for 5 subjects 
(9 videos per subject) with available ground-truth which is formatted as 
(\textit{X\_pos, Y\_pos, depth, roll, yaw (or pan), pitch (or tilt)}).

For each frame of one video sequence, we use the estimation of the rotation error $e_{i}=[\theta_{i} -\hat{\theta}_{i}]^{T}[\theta_{i} -\hat{\theta}_{i}]$ like \cite{lefevre-icme-2009} to evaluate the accuracy and robustness, where $\theta_{i}$ and $\hat{\theta}_{i}$ are $(pan, tilt, roll)$ of the ground-truth and estimated pose at frame $i$ respectively. A frame is lost when $e_{i}$ exceeds the threshold. The robustness is the number $N_{s}$ of frames tracked successfully and $P_{s}$ is the percentage of frames tracked over all videos. The precision measures include Pan, Tilt, Roll and average rotation errors which are computed by Mean Absolute Error (MAE) as the measure of tracker accuracy over tracked frames:
$E_{pan},E_{tilt},E_{roll}$ and $E_{m}=\frac{1}{3}\left(E_{pan}+E_{tilt}+E_{roll}\right)$ where $E_{pan}=\frac{1}{N_{s}}\sum_{i\in S_{s}}|\theta_{pan}^{i}-\hat{\theta}_{pan}^{i}|$ (similarly for the tilt and roll) and $S_{s}$ is set of tracked frames.

\textbf{The Talking Face Video:} is a freely 5000-frames video sequence of a talking person with face animations. The ground-truth is available with 68 facial points annotated manually on the whole video. Basing on movements of landmarks, we can estimate the face animation. On that account,  we instead evaluate the precision of landmark tracking as the accurate animation. The Root-Mean-Squared (RMS) error is normally used to evaluate the landmark tracking performance on this database. Despite that the number of landmarks of our system and other methods is different, the same evaluation scheme could be still applied on same number of landmarks with our work as well as other comparative methods.

\begin{table*}
\begin{centering}

\caption{The comparison of robustness ($P_{s}$) and accuracy ($E_{pan}$, $E_{tilt}$, $E_{roll}$ and $E_{avg}$) between our method and state-of-the-art on uniform-light set of BUFT dataset.}

\begin{tabular}{c c c c c c}
\hline 
Approach & $P_{s} $ & $E_{pan}$ & $E_{tilt}$ & $E_{roll}$ & $E_{avg}$\tabularnewline
\hline 
\hline 
(La Casicia \textit{et al.}, 2000) \cite{cascia-ieeepami-2000} & 75\% & 5.3 & 5.6 & 3.8 & 3.9 \tabularnewline
\hline 
(Xiao \textit{et al.}, 2003) \cite{xiao-ijist-2003} & 100\% & 3.8 & 3.2 & 1.4 & 2.8 \tabularnewline
\hline 
(Lefevre \textit{et al.}, 2009) \cite{lefevre-icme-2009} & 100\% & 4.4 & 3.3 & 2.0 & 3.2 \tabularnewline
\hline 
(Morency \textit{et al.}, 2008) \cite{morency-fg-2008} & 100\% & 5.0 & 3.7 & 2.9 & 3.9 \tabularnewline
\hline 
(Saragih \textit{et al.}, 2011) \cite{saragih-ijcv-2011} & 100\% & 4.3 $\pm$ 2.2 & 4.8 $\pm$ 3.3 & 2.6 $\pm$ 1.4 & 3.9 \tabularnewline
\hline 
(Chen \textit{et al.}, 2006) \cite{chen-bmvc-2006} & 91\% & 5.5 $\pm$ 1.7 & 4.2 $\pm$ 1.5 & 2.1 $\pm$ 1.0 & 3.9 \tabularnewline
\hline 
\textbf{Our method} & \textbf{100\%} & \textbf{5.4} $\pm$ \textbf{2.2} & \textbf{3.9} $\pm$ \textbf{1.7} & \textbf{2.4} $\pm$ \textbf{1.4} & 3.9 \tabularnewline
\hline 
\end{tabular}

\par\end{centering}

\end{table*}

The performance of pose estimation in Table 1 shows the comparable results between our work and state-of-the-art methods in 3d pose tracking. Our performance is $100\%$ robustness and the accuracy $E_{m}$ is 3.9, which outperforms \cite{chen-bmvc-2006} and \cite{cascia-ieeepami-2000} both in terms of robustness and accuracy. And it gets the same result of mean error $E_{m}$ as \cite{saragih-ijcv-2011,morency-fg-2008}, but the variance of error of \cite{saragih-ijcv-2011} is higher than our work especially in Tilt. However, we are worse than \cite{xiao-ijist-2003,lefevre-icme-2009} at the accuracy. In spite of the fact that our result is quite encouraging, the Pan precision is still low compared to others. The reason why Pan rotation is bad-estimated, could probably comes from occlusion problem. When Pan is bigger than, for instance, $30^{o}$, some landmarks are occluded that make local descriptors is inefficient that make the likelihood discontinued. For \cite{saragih-ijcv-2011}, the authors trained their landmarks classifiers only with variation of Pan angles that make their estimation of Tilt and Roll inefficient. Fig. 2 is an example of our method on one video of BUFT dataset.

In order to evaluate the landmark precision, we compare our method and FaceTracker\footnote{http://web.mac.com/jsaragih/FaceTracker/FaceTracker.html} proposed by \cite{saragih-ijcv-2011}. Because the landmarks of our method, \cite{saragih-ijcv-2011} and ground-truth are not the same, 12 landmarks around eyes, nose and mouth as in Fig. 3 are chosen to evaluate RMS error. The Fig. 4 shows the (Root Mean Square) RMS error which is computed using our method (red curve) and FaceTracker (blue curve) on the Talking Face video. The vertical axis is RMS error (in pixel) and the horizontal axis is the frame number. The model of \cite{saragih-ijcv-2011} sometimes drift away the ground-truth, but recovers quickly to good location by benefiting face and landmark detectors. The Fig. 4 shows that even though our method just learned from the synthesized database, what we obtain is the same the state-of-the-art method as well and is even more robust.

\begin{figure*}
\begin{centering}
\includegraphics[scale=0.60]{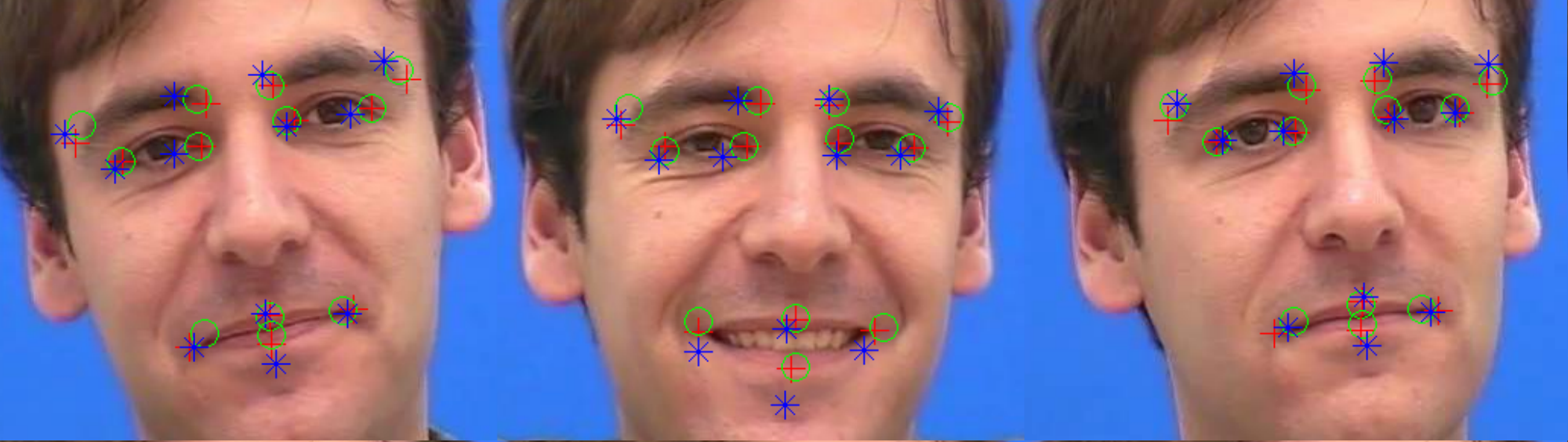}
\caption{The 12 landmarks is used to compute RMS error where red ($+$), blue ($*$) and green ($o$) markers are ground-truth, of Saragih \textit{et al.} \cite{saragih-ijcv-2011} and our method respectively on frames 110, 2500 and 4657 of Talking Face video.}
\end{centering}
\end{figure*}

The performance of our method for pose estimation could be improved if the Pan was estimated more accurately. One possible solution is assigning weights to landmarks corresponds to the Pan value. Or projecting landmarks on tangent plane at each landmark that compute mean and covariance matrices as a function of face pose to deal with occlusion. In general, how to deal with occluded landmarks is one of critical points to improve our performance. Although real-time computation is unreachable (about 5s/frame on Laptop Core 2 Duo 2.00GHz, 2G RAM) due to using down-hill simplex algorithm to optimize the energy function, it can be improved by using Gradient Descent in future work.


\begin{figure*}
\begin{centering}
\includegraphics[scale=0.60]{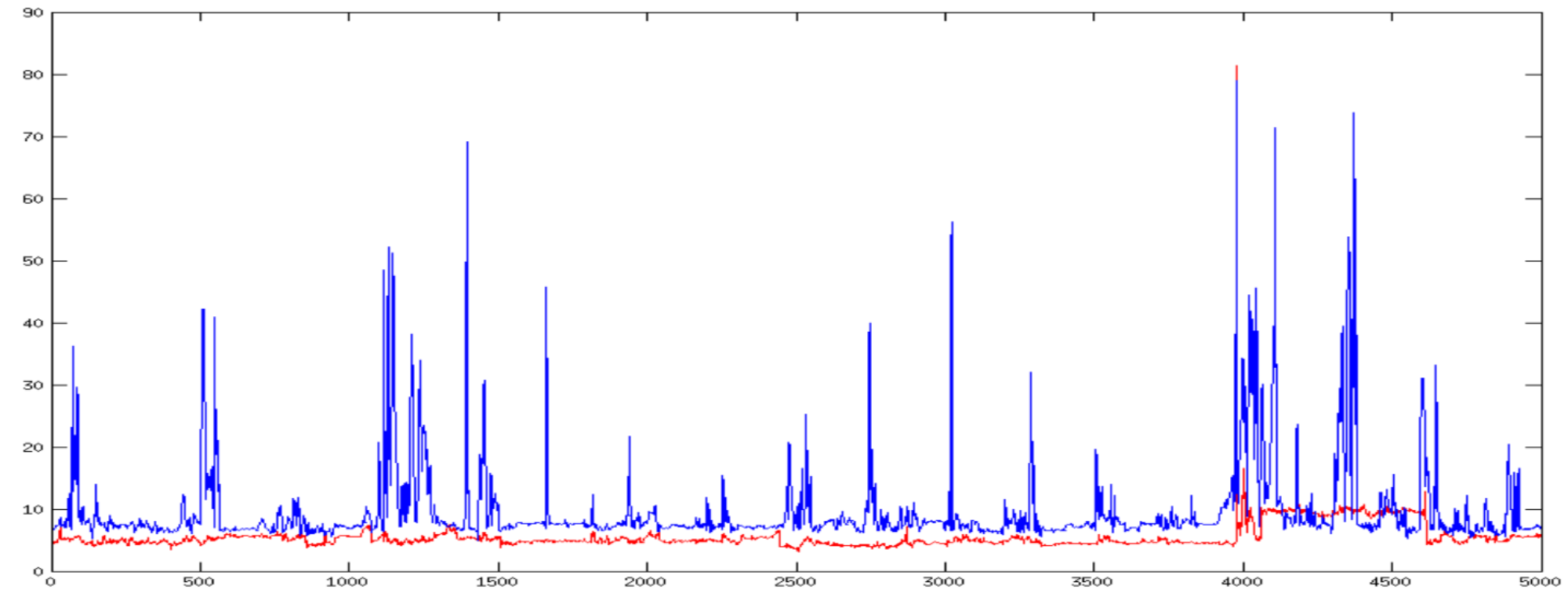}
\caption{The RMS error of 12 selected points for tracking in our framework (below red curve) and Saragih \textit{et al.} \cite{saragih-ijcv-2011} (above blue curve). The vertical axis is RMS error (in pixel) and the horizontal axis is the frame number.}.
\end{centering}
\end{figure*}


\section{Conclusion}
\label{sect:Word}

In this paper, we propose a Bayesian method to deal with the problem of face tracking using one adaptive model through eigen decomposition. The synthesized database within local features are around landmarks to learn appearance model as mean and covariance matrices. For tracking, an energy function which is approximated from posterior probability is minimized as difference between the observations and the appearance model. In order to adjust the model to changes of environments, the eigen decompostion is deployed. The results showed that the use of our model is comparable to some state-of-the-art methods of pose estimation and much more robust than state-of-the-art at landmark tracking or animation tracking. It demonstrated what we proposed is useful to both tasks of pose estimation and landmark tracking. Moreover, it is easy to build the learning database of synthesized images to learn without the need of real annotated data. With our current encouraging results, some other evolutions could be done to improve the performance. For examples, taking into account the weights of contribution to energy function which is dependent on the confidence of landmark observations at each time, computing appearance model as function of the pose to make the objective function continuous. In general, the way how to improve Pan precision by dealing with occluded landmarks is a crucial point to think as future work. Finally, the speed can be improved to real-time application by using Gradient Descent like methods instead of down-hill simplex algorithm.

\bibliographystyle{splncs}
\bibliography{isvc_submission}

\end{document}